\pdfoutput=1

\documentclass[11pt]{article}

\usepackage[]{ACL2023}

\usepackage{times}
\usepackage{latexsym}
\usepackage[utf8]{inputenc}
\usepackage{tcolorbox}
\usepackage{xcolor}
\usepackage{geometry}
\usepackage{microtype} 
\usepackage{placeins}
\usepackage{float}

\usepackage[T1]{fontenc}

\usepackage[utf8]{inputenc}

\usepackage{microtype}

\usepackage{inconsolata}

\title{When is dataset cartography ineffective? Using training dynamics does not improve robustness against Adversarial SQuAD}

\author{Paul K. Mandal\textsuperscript{\dag,$\diamondsuit$} \\
  \textsuperscript{\dag}Neurint LLC, Middletown, DE \\
  \textsuperscript{$\diamondsuit$}The University of Texas at Austin, Austin, TX \\
  \texttt{mandal@utexas.edu}
}

\begin{document}
\maketitle
\begin{abstract}
In this paper, I investigate the effectiveness of dataset cartography for extractive question answering on the SQuAD dataset. I begin by analyzing annotation artifacts in SQuAD and evaluate the impact of two adversarial datasets, AddSent and AddOneSent, on an ELECTRA-small model. Using training dynamics, I partition SQuAD into easy-to-learn, ambiguous, and hard-to-learn subsets. I then compare the performance of models trained on these subsets to those trained on randomly selected samples of equal size. Results show that training on cartography-based subsets does not improve generalization to the SQuAD validation set or the AddSent adversarial set. While the hard-to-learn subset yields a slightly higher F1 score on the AddOneSent dataset, the overall gains are limited. These findings suggest that dataset cartography provides little benefit for adversarial robustness in SQuAD-style QA tasks. I conclude by comparing these results to prior findings on SNLI and discuss possible reasons for the observed differences.
\end{abstract}

\section{Introduction}

Human-labeled datasets, such as the Stanford Natural Language Inference (SNLI) corpus \citep{snli}, often introduce biases that are not representative of naturally occurring data. These biases can manifest as annotation artifacts, allowing models to exploit spurious patterns rather than genuinely understanding the underlying task \cite{mccoy-etal-2019-right}. As a result, training on datasets like SNLI can lead to artificially inflated accuracies that exceed what could reasonably be expected in real-world scenarios.

\citet{gururangan2018annotation} highlights the extent of such artifacts in SNLI, achieving 67\% accuracy using a hypothesis-only model. This finding underscores the presence of significant annotation artifacts, as the model did not require information from the premise to predict labels accurately. Additionally, \citet{feng2019misleading} demonstrates that the absence of high performance in hypothesis-only models does not guarantee the absence of artifacts. Their study identifies biases in SNLI that remain undetectable by partial-input models, suggesting that such methods may provide an incomplete picture of dataset biases.

In this paper, I conduct a similar analysis of the Stanford Question Answering Dataset (SQuAD) \citeyearpar{squad}. I first evaluate the performance of an ELECTRA-small model \citeyearpar{electra} trained on SQuAD on two adversarial datasets proposed in \citet{jia-adversarial}. I subsequently leverage dataset cartography to categorize training examples into easy-to-learn, ambiguous, and hard-to-learn subsets based on their training dynamics as proposed by \citet{dataset_cartography}. Furthermore, I evaluate the performance of an ELECTRA-small model \citeyearpar{electra} trained on these three subsets to assess how dataset difficulty influences model behavior on both regular SQuAD and adversarial SQuAD. Finally, I interpret these findings in my discussion.

\section{Adversarial Examples}
\citet{jia-adversarial} introduced an adversarial evaluation framework for the Stanford Question Answering Dataset (SQuAD). This adversarial framework demonstrates that existing models are highly susceptible to these adversarial modifications, revealing significant weaknesses in their ability to generalize beyond surface-level patterns. 

\newcommand{\highlightgreen}[1]{\textcolor{green!70!black}{#1}}
\newcommand{\highlightred}[1]{\textcolor{red!70!black}{#1}}
\newcommand{\highlightblue}[1]{\textcolor{blue}{#1}}

\begin{figure*}[htbp]
\centering
\begin{tcolorbox}[colframe=black!50,colback=white,sharp corners]
\textbf{ID:} 56bf10f43aeaaa14008c94fd \\
\textbf{Title:} Super\_Bowl\_50 \\[0.10em]

\textbf{Context:} \\
Super Bowl 50 was an American football game to determine the champion of the National Football League (NFL) for the 2015 season. The
American Football Conference (AFC) champion Denver Broncos defeated the National Football Conference (NFC) champion Carolina Panthers 24–10 to earn their third Super Bowl title. The game was played on February 7,
2016, at Levi's Stadium in the San Francisco Bay Area at Santa Clara, California. As this was the 50th Super Bowl, the league emphasized the "golden anniversary" with various gold-themed initiatives, as well as temporarily suspending the tradition of naming each Super Bowl game with
Roman numerals (under which the game would have been known as "Super Bowl L"), so that the logo could prominently feature the Arabic numerals 50. \highlightblue{The Dallas Buccaneers secured a Champ Bowl title for the third time in the year of 1990.} \\[0.10em]

\textbf{Question:} What year did the Denver Broncos secure a Super Bowl title for the third time?

\textbf{Correct Answers:} 2015, 2016\\[0.10em]

\textbf{File 1 Predicted Answer:} \highlightgreen{2016} \\
\textbf{File 2 Predicted Answer:} \highlightred{1990}
\end{tcolorbox}
\caption{Comparison of predicted answers on original and AddSent adversarial contexts using the ELECTRA-small model. Correct predictions are green, incorrect predictions are red, and the adversarial sentence is shown in blue.}
\label{fig:addSent}
\end{figure*}

\citet{jia-adversarial} introduces two adversarial evaluation datasets, \textbf{AddSent} and \textbf{AddOneSent}, which modify the SQuAD validation set to test model robustness. \textbf{AddSent} appends grammatically correct and semantically plausible but misleading sentences 
designed to confuse the model by mimicking answer patterns. \textbf{AddOneSent} selects a random human approved sentence and appends it to the context. I use both datasets to evaluate the robustness of our baseline model.

\section{A Primer on Training Dynamics and Dataset Cartography}

\citet{dataset_cartography} was the seminal paper that proposed training dynamics and dataset cartography.

\textbf{Training Dynamics} involve the temporal progression of model behavior as it learns from data. They are characterized by how a model's predictions, confidence scores, and gradients evolve over epochs. Training dynamics not only reveal insights about the convergence properties of the model but also highlight systemic biases or inconsistencies in the dataset.

\textbf{Dataset Cartography} refers to the process of analyzing and categorizing training examples based on their learning dynamics during model training. This approach provides insights into the complexity and informativeness of individual examples within a dataset, helping to identify patterns such as hard-to-learn or ambiguous examples. By monitoring metrics like confidence, variability, and correctness, dataset cartography constructs a three-dimensional map of the training data. This map divides examples into regions such as \textit{easy-to-learn}, \textit{ambiguous}, or \textit{hard-to-learn}.

\begin{figure}[t]
    \centering
    \includegraphics[width=1\linewidth]{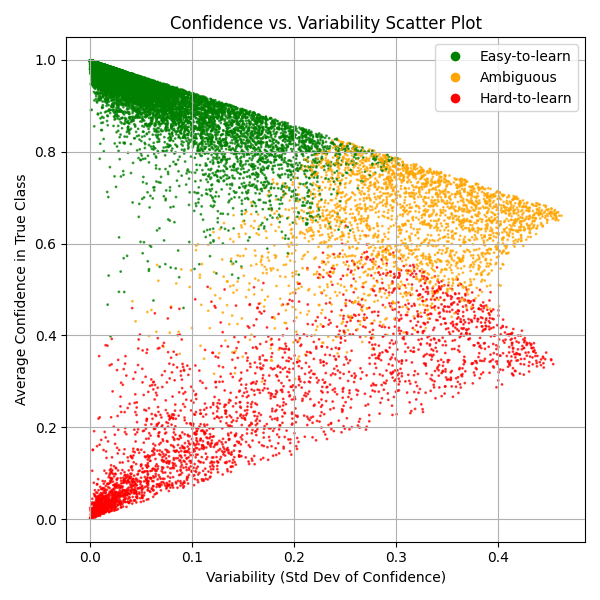}
    \caption{Data map for 10K random examples from the SNLI train set on an ELECTRA-small classifier. The $x$-axis shows variability while the $y$-axis shows confidence. I use green, yellow, and red to illustrate the easy-to-learn, ambiguous, and hard-to-learn training samples respectively.}
    \label{fig:snli-dynamics}
\end{figure}

\begin{figure}[t]
    \centering
    \includegraphics[width=1\linewidth]{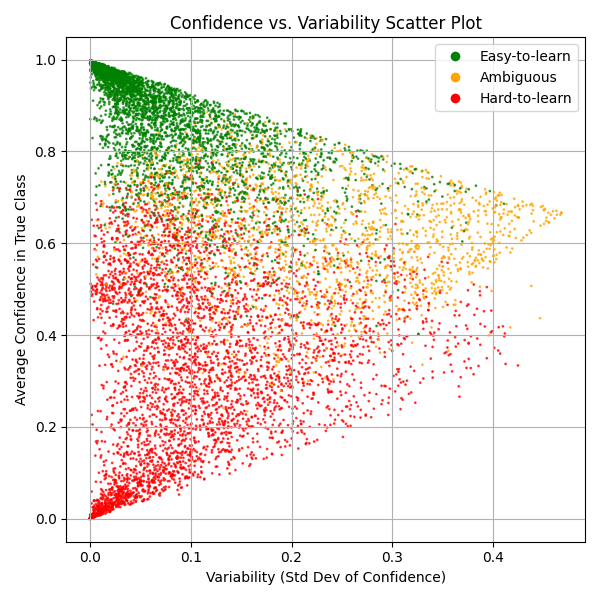}
    \caption{Data map for 10K random examples from the SQuAD train set on an ELECTRA-small classifier. The $x$-axis shows variability while the $y$-axis shows confidence. I use green, yellow, and red to illustrate the easy-to-learn, ambiguous, and hard-to-learn training samples respectively.}
    \label{fig:squad-dynamics}
\end{figure}

\section{Analysis of Baseline Model}
\label{sec:analysis}

To evaluate the robustness of the baseline ELECTRA model, I trained it on both the SQuAD and SNLI datasets, leveraging the dataset cartography framework to analyze training dynamics. This approach allowed for a systematic categorization of data points into easy-to-learn, ambiguous, and hard-to-learn subsets.

To further challenge the model, I evaluated the baseline ELECTRA-small model against the AddSent and AddOneSent adversarial datasets introduced by \citet{jia-adversarial}. These adversarial examples are designed to mislead question-answering systems by introducing semantically plausible distractors. Other work has similarly found that extractive question answering models often rely heavily on superficial cues such as lexical overlap and sentence position rather than deep reasoning \cite{sugawara-etal-2018-makes}. As shown in Tables \ref{tab:baseline_addsent_eval} and \ref{tab:baseline_addonesent_eval}, the model experienced a notable drop in performance on both datasets, indicating a vulnerability to adversarial perturbations despite strong performance on the original SQuAD validation set. 

An adversarial example from the \textbf{AddOne} dataset is shown in Figure \ref{fig:addSent}. The adversarial sentences, highlighted in blue, introduced plausible but misleading distractors to test the model's robustness. In the example, the model incorrectly selects "1990" over the correct answer, "2016," due to the adversarial sentence's proximity and plausibility. These results highlight vulnerabilities in handling distractors, emphasizing the need for improved comprehension and prioritization in question-answering systems.

Figures \ref{fig:snli-dynamics} and \ref{fig:squad-dynamics} illustrate the training dynamics of the SQuAD and SNLI datasets, respectively. As evident from these plots, the training dynamics of the SQuAD dataset exhibit significantly more variance compared to SNLI. This variance suggests that SQuAD contains a wider range of difficulty levels or inherent noise, potentially contributing to a more complex learning process for the model.

\begin{table}[t]
\centering
\begin{tabular}{lcc}
\hline
 & \textbf{Exact Match} & \textbf{F1} \\
\hline
Original & 76.91 & 85.28 \\
AddSent & 53.74 & 60.70 \\
\hline
\end{tabular}
\caption{Performance of baseline model on AddSent adversarial set compared to the original examples}
\label{tab:baseline_addsent_eval}
\end{table}

\begin{table}[t]
\centering
\begin{tabular}{lcc}
\hline
 & \textbf{Exact Match} & \textbf{F1} \\
\hline
Original & 77.28 & 85.14 \\
AddOneSent & 63.07 & 70.31 \\
\hline
\end{tabular}
\caption{Performance of baseline model on AddOneSent adversarial set compared to the original examples}
\label{tab:baseline_addonesent_eval}
\end{table}

\section{Results}

To investigate the effectiveness of dataset cartography for mitigating artifacts in SQuAD, I divided the dataset into three subsets: easy-to-learn, ambiguous, and hard-to-learn examples. These subsets were determined based on training dynamics, which track metrics such as prediction confidence and variability over training epochs. An ELECTRA-small model was trained separately on each subset to assess the impact of these divisions on model performance. Additionally, a model was trained on 33\% of the original dataset to serve as a baseline, as the size of each subset was approximately one-third of the total training data.

The evaluation was performed on three datasets: the SQuAD validation set, the AddSent adversarial set, and the AddOneSent adversarial set. The results revealed that dataset cartography offered limited benefits for SQuAD compared to its previously demonstrated effectiveness with SNLI. The model trained on the hard-to-learn subset showed a marginally better F1 score on the AddOneSent set, indicating some improvement in robustness to adversarial distractors. However, for both the SQuAD validation set and the AddSent adversarial set, the models trained on easy-to-learn, ambiguous, and hard-to-learn subsets all performed similarly to the baseline model trained on 33\% of the data. This suggests that while dataset cartography can influence robustness in specific cases, its overall impact on SQuAD is minimal.

\begin{table}
\centering
\begin{tabular}{lcc}
\hline
 & \textbf{Exact Match} & \textbf{F1} \\
\hline
100\% train (Baseline) & 78.33 & 86.09 \\
\hline
33\% train & \textbf{73.67} & \textbf{82.70} \\
easy-to-train & 72.97 & 81.03 \\
ambiguous & 73.15 & 81.59 \\
hard-to-train & 67.28 & 80.27 \\
\hline
\end{tabular}
\caption{Results on the SQuAD validation set.}
\label{tab:accents}
\end{table}

\begin{table}
\centering
\begin{tabular}{lcc}
\hline
 & \textbf{Exact Match} & \textbf{F1} \\
\hline
100\% train (Baseline) & 53.73 & 60.70 \\
\hline
33\% train & \textbf{48.03} & 55.14 \\
easy-to-train & 46.57 & 52.34 \\
ambiguous & 45.65 & 51.93 \\
hard-to-train & 45.73 & \textbf{56.56} \\
\hline
\end{tabular}
\caption{Results on the AddOneSent adversarial set.}
\label{tab:accents}
\end{table}

\begin{table}[t]
\centering
\begin{tabular}{lcc}
\hline
 & \textbf{Exact Match} & \textbf{F1} \\
\hline
100\% train (Baseline) & 63.07 & 70.31 \\
\hline
33\% train & \textbf{58.37} & \textbf{65.99} \\
easy-to-train & 56.59 & 63.44 \\
ambiguous & 56.58 & 63.50 \\
hard-to-train & 53.72 & 65.47 \\
\hline
\end{tabular}
\caption{Results on the AddSent adversarial set.}
\label{tab:accents}
\end{table}

\section{Discussion}
\label{sec:discussion}

The findings highlight two primary reasons for the reduced effectiveness of dataset cartography on SQuAD compared to SNLI.

First, the nature of artifacts in SQuAD is less straightforward than those in SNLI. In SNLI, artifacts arise from human-generated annotations, which often introduce predictable patterns, especially in constructing contradictions. These patterns make it easier for dataset cartography to identify and mitigate artifacts. For example, hypotheses in SNLI labeled as contradictions frequently include explicit negations, allowing models to rely on shallow heuristics that dataset cartography can address effectively. In contrast, SQuAD artifacts are more nuanced and embedded within the task of predicting exact answers from a passage. This difference makes it more challenging for dataset cartography to isolate problematic examples that contribute to model errors.

Second, SQuAD's higher variance in training dynamics further complicates the application of dataset. The SQuAD dataset requires models to extract precise answers from diverse and often complex contexts, as opposed to SNLI's simpler task of classification into one of three categories: entailment, contradiction, or neutral making it more susceptible to breakdowns when small semantic or syntactic changes are introduced \cite{gardner-etal-2020-evaluating}. This broader problem scope introduces variability in the learnability of examples, with some requiring detailed reasoning. While dataset cartography can identify subsets based on training behavior, this variance reduces the distinction between easy-to-learn, ambiguous, and hard-to-learn subsets in terms of their overall impact on model performance.

Additionally, the marginal improvement observed for the hard-to-learn subset on the AddOneSent evaluation suggests that this subset may be better aligned with the adversarial task of handling distractors. However, the absence of significant differences across the other evaluations indicates that the subsets identified by dataset cartography are not sufficiently distinct in SQuAD to substantially influence generalization or robustness.

These results underscore the limitations of applying dataset cartography uniformly across different datasets and tasks. While it has proven effective for simpler classification tasks like SNLI, its utility for more complex datasets like SQuAD is less clear. Future research could explore adaptations of dataset cartography to address datasets with high variance or tasks requiring nuanced reasoning, such as question answering. My implementation is publicly available as a resource for future work. \footnote{\url{https://github.com/PaulKMandal/dataset-artifacts}} 

\section*{Limitations}
Unlike \citet{dataset_cartography}, I did not train our models multiple times to calculate the standard deviation of our model performance. 
\section*{Ethics Statement}
There are no major ethical considerations in this work.

\bibliography{anthology,custom}
\bibliographystyle{acl_natbib}

\appendix



\end{document}